\documentclass[acmsmall]{acmart}

\usepackage[ruled,linesnumbered]{algorithm2e}
\usepackage{multirow}
\usepackage{makecell}
\usepackage{diagbox}
\usepackage{float}
\usepackage{xcolor}
\usepackage{hyperref}
\usepackage{subcaption}

\definecolor{brown}{rgb}{0.59, 0.29, 0.0}
\definecolor{orange}{rgb}{0.93, 0.53, 0.18}

\newcommand{\hl}{}
\newcommand{\ie}{\emph{i.e.}}
\newcommand{\eg}{\emph{e.g.}}
\newcommand{\etc}{etc}

\AtBeginDocument{%
  }

\setcopyright{acmlicensed}
\copyrightyear{2018}
\acmYear{2018}
\acmDOI{XXXXXXX.XXXXXXX}

\acmJournal{JACM}
\acmVolume{37}
\acmNumber{4}
\acmArticle{111}
\acmMonth{8}

\begin{document}

\title{SOEDiff: Efficient Distillation for Small Object Editing}

\author{Yiming Wu}
\email{yimingwu0@gmail.com}
\authornote{First two authors contributed equally to this research.}
\orcid{0000-0002-9866-669X}
\affiliation{
  \institution{The University of Hong Kong}
  \city{Hong Kong}
  \state{Hong Kong SAR}
  \country{China}
}

\author{Qihe Pan}
\email{panqihe996@gmail.com}
\authornotemark[1]
\affiliation{
  \institution{Zhejiang University of Technology}
  \city{Hangzhou}
  \state{Zhejiang}
  \country{China}
}

\author{Zhen Zhao}
\email{zhen.zhao@sydney.edu.au}
\affiliation{
  \institution{The University of Sydney}
  \city{Sydney}
  \state{NSW}
  \country{Australia}
}

\author{Zicheng Wang}
\email{xiaoyao3302@outlook.com}
\affiliation{
  \institution{The University of Hong Kong}
  \city{Hong Kong}
  \state{Hong Kong SAR}
  \country{China}
}

\author{Sifan Long}
\email{longsf22@mails.jlu.edu.cn}
\affiliation{
  \institution{Jilin University}
  \city{Changchun}
  \state{Jilin}
  \country{China}
}

\author{Ronghua Liang}
\email{rhliang@zjut.edu.cn}
\affiliation{
  \institution{Zhejiang University of Technology}
  \city{Hangzhou}
  \state{Zhejiang}
  \country{China}
}

\renewcommand{\shortauthors}{Wu et al.}
\begin{abstract}
In this paper, we delve into a new task known as small object editing (SOE), which focuses on text-based image inpainting within a constrained, small-sized area. Despite the remarkable success have been achieved by current image inpainting approaches, their application to the SOE task generally results in failure cases such as \textit{Object Missing, Text-Image Mismatch, and Distortion}. These failures stem from the limited use of small-sized objects in training datasets and the downsampling operations employed by U-Net models, which hinders accurate generation. To overcome these challenges, we introduce a novel training-based approach, SOEDiff, aimed at enhancing the capability of baseline models like StableDiffusion in editing small-sized objects while minimizing training costs. Specifically, our method involves two key components: \textbf{SO-LoRA}, which efficiently fine-tunes low-rank matrices, and \textbf{Cross-scale score distillation}, which leverages high-resolution predictions from the pre-trained teacher diffusion model. Our method presents significant improvements on the test dataset collected from MSCOCO and OpenImage, validating the effectiveness of our proposed method in small object editing. In particular, when comparing SOEDiff with SD-I model on the \textit{OpenImage-small-val} dataset, we observe a 0.99 improvement in CLIP-Score and a reduction of 2.87 in FID.
\end{abstract}



\received{20 February 2007}
\received[revised]{12 March 2009}
\received[accepted]{5 June 2009}

\maketitle

\begin{figure}
\centering
  \includegraphics[width=0.99\linewidth]{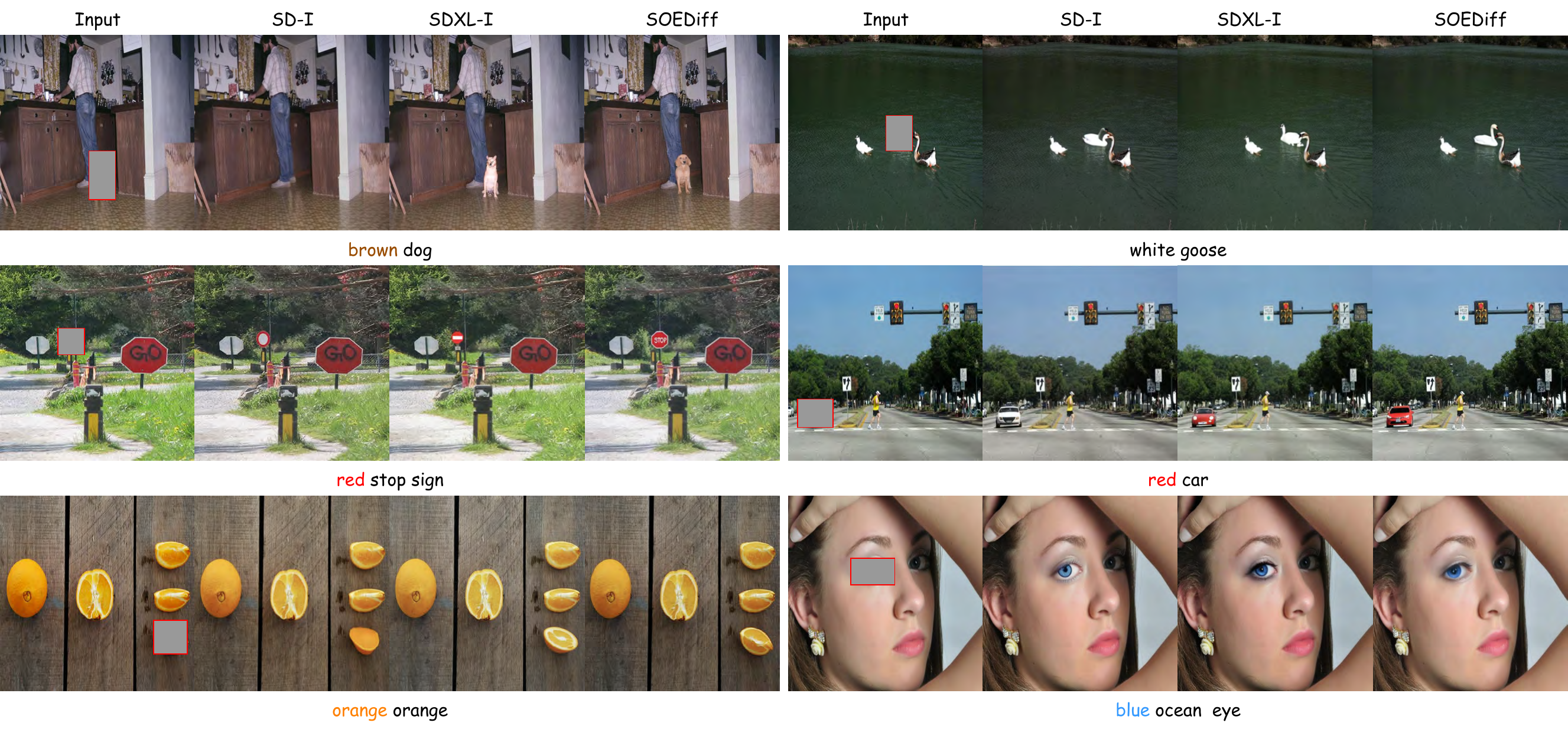}
  \caption{The image showcases examples of small object editing. The first column shows input images along with the masked small areas highlighted within red bounding boxes. The second column depicts results generated by SD-I~\cite{rombach2022high}. In the third column, results produced by SD-XL~\cite{podell2023sdxl} are presented. The fourth column features outcomes generated by our proposed SOEDiff. ``\textcolor{brown}{brown}, \textcolor{red}{red}, \textcolor{orange}{orange}, white, and \textcolor{blue}{blue}'' are colors for editing objects (\ie, dog, goose, stop sign, \etc). To have a better view, please visit our project page \url{https://soediff.github.io}.}
  \Description{}
  \label{fig:teaser}
\end{figure}

\section{Introduction}~\label{sec:introduction}
The advent of diffusion models has marked a new era in the field of text-to-image generation, demonstrating the creation of intricate and coherent visual representations. Compared with generating entirely new images, image-editing models present particular excellence in manipulating regional aspects adjustment (\textit{e.g.,} the appearance and objects) and holistic style transformation (\textit{e.g.,} the structure, aesthetic, and style) of images. Despite the progress achieved by the recent researcher in generic-sized object editing, there has been a large oversight in addressing the requirement for subtle adjustments and fine-grained editing in extremely small areas (\textit{e.g.,} changing the style of eyebrow or removing the barrettes). Given this perspective, we pose the question naturally: \textit{What happens if we perform image editing in small areas?}

\begin{figure}[t]
  \centering
  \begin{subfigure}[b]{0.44\textwidth}
  \centering
    \includegraphics[width=1\linewidth]{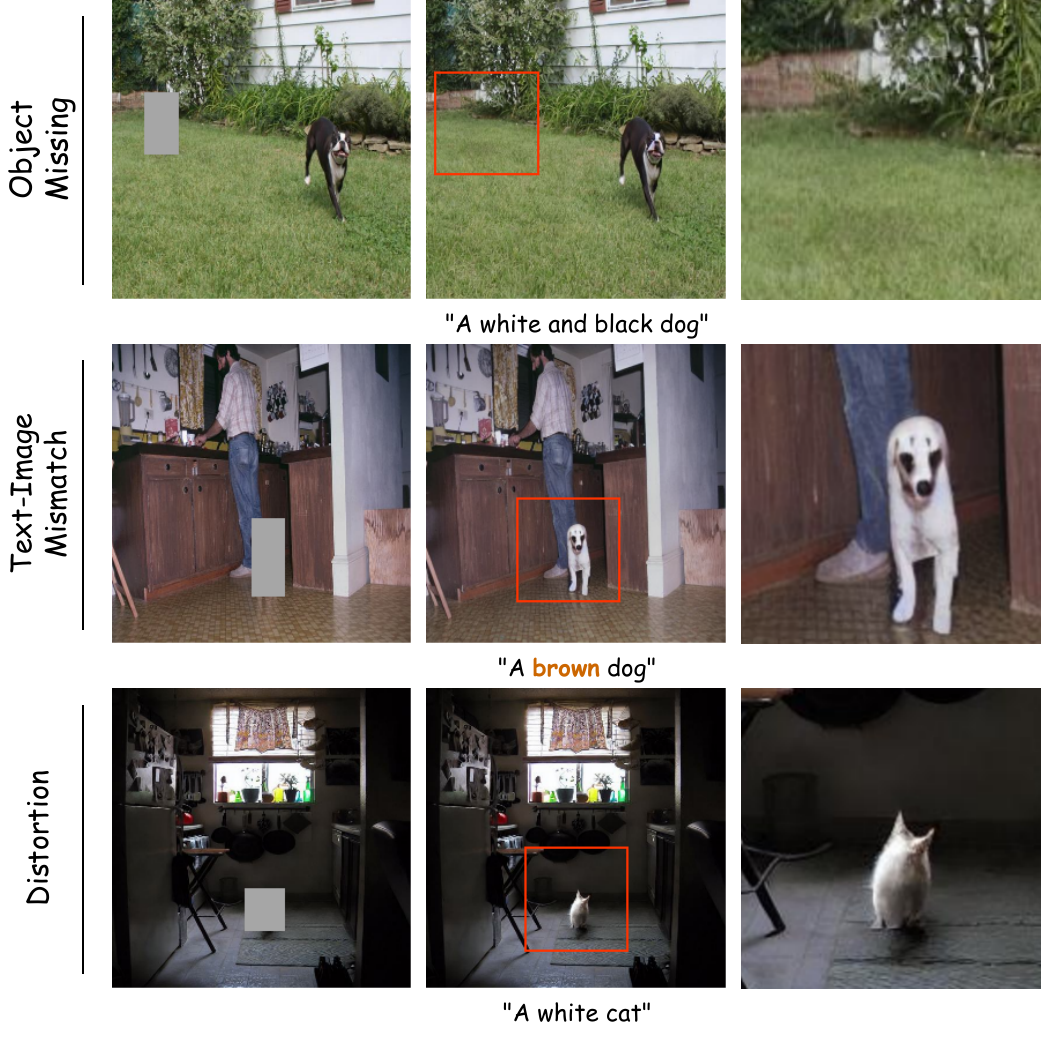}
    \caption{}
    \label{fig:challenges}
  \end{subfigure}
  \begin{subfigure}[b]{0.55\textwidth}  
    \includegraphics[width=1\linewidth]{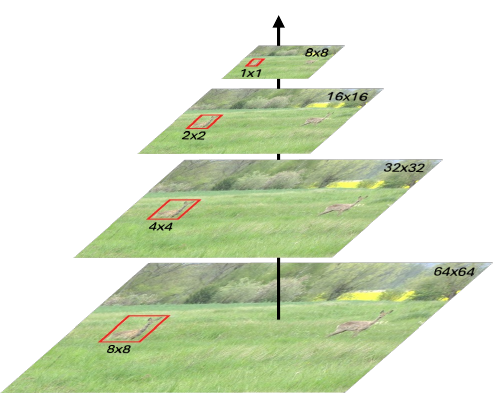}
    \caption{}
    \label{fig:analysis}
  \end{subfigure}
  \caption{(a) Challenges in editing small objects. Images are presented along with their corresponding masks, edited images, and enlarged areas in three separate columns. Three major challenges are presented: \textbf{Object Missing,} the model fails to generate the object as described in the text description; \textbf{T2I Mismatch,} the discrepancy between the generated content and the textual description, particularly in attributions like color or shape; \textbf{Distortion,} the generated object appears distorted, \textit{e.g.}, the essential features like the cat's face are missing in this image. The images are generated by SD-I. (b) The illustration of a cross-attention map. For an input image with a size of $512 \times 512$, if the masked area is $64 \times 64$, the corresponding effective area comes to $1 \times 1$ in the mid-block. The diminutive size of this masked area poses a challenge as it may lack sufficient semantic information essential for generating associated objects. Zoom in for a better view.}
  \Description{}
\end{figure}

The constraints of editing in a confined small area pose significant challenges to conventional editing techniques, which we refer to as Small Object Editing (SOE)~\footnote{In this paper, our focus is on the task of inpainting the object within a small-sized bounding box, which are subsequently used to create mask.}.
As shown in Fig.~\ref{fig:challenges}, we present three typical types of failure cases: \emph{Object Missing}, \emph{Text-Image Mismatch}, and \emph{Distortion}. 
These failures stem from two main factors.
First, small-sized objects are rarely used in training image editing models, thereby restricting the capability to generate small objects. For instance, GLIGEN~\cite{li23gligen} discards the small-sized objects during the pre-processing of the training datasets. Consequently, lacking such training information can severely limit the ability to follow specific editing instructions on small objects and even lead to image distortion around the objects.
Second, upon inspecting the architecture of the convolutional U-Net models, we find that the down-sampling operation employed by StableDiffusion~\cite{rombach2022high} impedes the accurate generation of small objects. Specifically, we denote the grids corresponding to the original bounding box in the feature maps as ``effective area'', which diminishes in size due to multiple down-sampling operations. As illustrated in Fig.~\ref{fig:analysis}, the size of the cross-attention map progressively diminishes as the depth increases. At the mid-block (depicted in the top parallelogram), the size of the cross attention map is $8\times8$, with each side length being 1/8 of the original latent code $z$. At this juncture, the corresponding feature map size dwindles to $1\times1$. The diminutive size of this masked area poses a challenge as it may lack sufficient semantic information essential for generating associated objects. This challenge intensifies the difficulty of accurately injecting the textual information into the corresponding ``effective area'' in the cross-attention layer.

To overcome the challenges, we propose a novel method termed SOEDiff, designed to improve the capabilities of base models (\textit{e.g.}, StableDiffusion) in editing small-sized objects while minimizing training costs. To this end, two main components are involved in our approach: First, drawing inspiration from LoRA~\cite{hu2021lora} which is known for its efficiency in fine-tuning diffusion models, we propose the SO-LoRA, which aims to optimize low-rank matrices for small object editing, enhancing the alignment between textual descriptions and generated small objects. Second, we propose cross-scale score distillation to further enhance the image quality. By leveraging insights from high-resolution predictions through a pre-trained teacher diffusion model, we mitigate blurriness and distortion typical in traditional diffusion models. Additionally, to further enhance the image fidelity, we fine-tune the VAE by performing pixel-level reconstruction.
Our main contributions are as follows:

\begin{itemize}
    \item \hl{We introduce a novel task termed small object editing (SOE) and establish a comprehensive benchmark dataset~\footnote{dataset is available at \url{https://soediff.github.io/}} for evaluating SOE performance. This dataset comprises 300K images for training and two 2K image sets for testing. Our analysis reveals that existing image generation models, such as SD1.5 and SDXL, exhibit significant performance degradation on the SOE task.}
    \item \hl{To address these challenges, we propose a new method, \textbf{SOEDiff}, which incorporates a cross-scale distillation mechanism to manage the scale variation of editing objects. This mechanism leverages high-resolution teacher representations to guide the low-resolution student model. Additionally, we employ low-rank adaptation (LoRA) to fine-tune the student model, resulting in a lightweight and efficient model for SOE.}
    \item \hl{We conduct extensive experiments on the proposed SOE benchmark dataset to evaluate the performance of SOEDiff. The results demonstrate that SOEDiff significantly outperforms existing methods in terms of editing quality, diversity, and efficiency. We also provide detailed analyses and ablation studies to validate the effectiveness of our approach.}

\end{itemize}

\section{Related Work}

\subsection{Text-to-Image Diffusion}
Diffusion models (DMs)~\cite{podell2023sdxl, rombach2022high, ramesh2022hierarchical, nichol2021glide, ho2020denoising, wang2024mix} have emerged as the \emph{de facto} standard in the domain of image generation, outperforming Generative Adversarial Networks (GANs)~\cite{goodfellow2014generative, tao2022df, zhang2021cross, zhu2019dm, xu2018attngan} in terms of both superior performance and training robustness. Denoising Diffusion Probabilistic Models (DDPM)~\cite{ho2020denoising} and Noise Conditional Score Networks (NCSN)~\cite{song2019generative} pioneered the utilization of denoising neural networks to invert a predefined Markovian noising process applied to raw images. Latent Diffusion Models (LDM)~\cite{rombach2022high} intuitively disentangle image synthesis into two stages: semantic reconstruction through a denoising process in latent space, and perceptual reconstruction through decoding, such a strategy significantly enhances visual fidelity. Further advancing the design of effective network architectures, Imagen~\cite{saharia2022photorealistic} integrates the capabilities of a pre-trained large language model (\textit{e.g.}, T5~\cite{raffel2020exploring}) to enhance image-text alignment and introduces additional diffusion models for super-resolution, thereby generating images of larger sizes. Similarly, SDXL~\cite{podell2023sdxl} extends the architecture by incorporating a larger denoising network and an extra refiner network, improving visual fidelity. DALLE-3~\cite{betker2023improving} enhances the prompt-following capabilities of DMs by training on highly descriptive captions generated through a bespoke image captioning process applied to the training dataset. 
Those diffusion models are conditioned mainly on text prompts. ControlNet~\cite{zhang2023adding} and T2I-Adapter~\cite{mou2024t2i} indeed mark another significant milestone in the development of image-conditioned generation.
Moving beyond the UNet-like architecture, DiTs~\cite{peebles2023scalable} and U-Vit~\cite{bao2023all} have explored the replacement of UNet with transformer blocks, achieving superior performance by scaling up diffusion models.

Similar to the success in the image synthesis domain, diffusion has had significant strides in editing images too. 
Imagic~\cite{kawar2023imagic} is the first able to make complex non-rigid edits to real images by learning to align a text embedding with the input image and the target text. 
PnP~\cite{tumanyan2023plug}, NTI~\cite{mokady2023null}, Imagic~\cite{kawar2023imagic}, DiffEdit~\cite{couairon2022diffedit} first use DDIM~\cite{song2020denoising} inversion to invert the image to the input tensor, and then use the text conditioned denoising process to generate the image with the required edit. 
P2P~\cite{hertz2022prompt} achieves modifications to synthesized images by utilizing attention control.
InstructPix2Pix~\cite{brooks2023instructpix2pix} and Imagen Editor~\cite{saharia2022photorealistic}  pass the image to be edited directly to the diffusion model by bypassing the DDIM inversion step. 
DiffusionCLIP~\cite{kim2022diffusionclip} leveraged the capability of CLIP~\cite{radford2021learning} to perform image-text feature alignment to achieve the text-driven editing of the image. 
BlendedDiffusion~\cite{avrahami2022blended, avrahami2023blended} explore the method of using a mask to edit the specific region and add a new object to the image while leaving the rest unchanged, while ESD~\cite{gandikota2023erasing} and Inst-Inpaint~\cite{yildirim2023inst} focus on erasing the concept in the image by instructions. Recently, multi-object editing approaches have been explored. LEDITS++~\cite{brack2024ledits++} proposes an inversion-based approach that supports multiple simultaneous edits without tuning or optimization. AnyDoor~\cite{chen2024anydoor} considers zero-shot multi-object editing which could teleport multiple unseen objects to a scene with the desired location and shape via disentangling identity information and detail high-frequency map from the reference image. Similarly, LoMOE~\cite{chakrabarty2024lomoe} proposes MultiDiffusion to handle the attribute preservation and multi-object editing simultaneously, and release a dataset dedicated to multi-object editing named LoMOE-Bench.
LGG~\cite{pan2024towards} \hl{is a pioneering work that focuses on the Small Object Editing (SOE) task, establishing a benchmark for evaluating generative models' performance in editing small objects or regions within images. Unlike the training-free approach employed in LGG, our work introduces a novel training-based method, enhancing model performance by fine-tuning the pre-trained model with a newly curated dataset of 300K images.}

\subsection{Parameter-efficient Fine-tuning}
Parameter-efficient fine-tuning (PEFT) concentrates on fine-tuning and optimizing model parameters in a way that conserves resources. This strategy is designed to boost a pre-trained model's efficacy for particular tasks or areas, avoiding the need for substantial extra data or computational power. Through the targeted adjustment of a select group of parameters, PEFT effectively adapts the model, proving to be highly beneficial in situations where labeled data is scarce or computational resources are restricted.
Among many PEFT methods, the method based on reparameterization is the most representative and has been widely researched. LoRA~\cite{hu2021lora} leverages low-rank approximations to efficiently fine-tune the model parameters, enabling more effective adaptation while mitigating computational demands. Specifically tailored for Stable Diffusion models, LoRA offers a promising avenue to enhance adaptability and performance in generative tasks involving diffusion processes. ControlNet~\cite{zhang2023adding} and T2I-Adapter~\cite{mou2024t2i} employ LoRA to process a variety of image prompts, enabling broad control over image characteristics. It allows various conditional inputs, such as edge and pose, to control the image generation process. The latent consistency model (LCM)~\cite{luo2023latent} leverages LoRA and distillation tech for model acceleration during the sampling process and achieved great success. 
ConceptSliders~\cite{gandikota2023concept} and ContinualDiffusion~\cite{smith2023continual} using LoRA to achieve more precise control.
At the same time, LoRA can also be used with Dreambooth~\cite{ruiz2023dreambooth} to help the model learn a new concept, which can be an object or a style.
\subsection{Diffusion Distillation}
Distillation technology is widely utilized for accelerating the sampling tasks in diffusion models. In the diffusion distillation framework, a student model is trained to distill the multi-step outputs of the original diffusion model into fewer steps. Progressive Distillation~\cite{salimans2022progressive} train a series of student models by repeated distillation. Adversarial Diffusion Distillation(ADD)~\cite{sauer2023adversarial}, uses a pre-trained feature extractor as its discriminator, achieving great performance in only four steps. Consistency Model(CM)~\cite{song2023consistency} employs a consistency mapping tech to achieve one-step generation. Inspired by CM, LCM~\cite{luo2023latent} solves the reverse diffusion process by treating it as a probability flow ODE (PF-ODE) problem in the latent space. And LCM-LoRA~\cite{luo2023lcm} introduces LoRA training for efficiently learning LCM modules. SD3-Turbo applies ADD to SD3~\cite{esser2024scaling} to achieve the goal of fast high-resolution image synthesis.

In our paper, we are motivated by existing work in the field, and we have designed a LoRA specifically aimed at addressing the issue of small object generation with distillation. Our method fills a critical gap in the current landscape, offering a more universal and accurate solution for this challenge. This advancement not only improves upon the existing methodologies but also broadens the scope of applicability in the domain of diffusion models for image generation.
\section{Preliminaries}\label{sec:preliminaries}
\hl{Diffusion model is a kind of generative model that reverses the diffusion process to synthesize data. Initially, the forward diffusion process gradually adds noise to the data, transitioning it from a data point \textit{$x_0$} to the complete Gaussian noise \textit{$x_T$}. At any timestep \textit{t}, the noised image is modeled as:}
\begin{equation}
\label{eq:diffusion process}
    x_t = \sqrt[]{1-\overline{\beta_t}}x_0 \ +\  \sqrt[]{\overline{\beta_t}} \epsilon,
\end{equation}
\hl{where $\epsilon$ is randomly sampled Gaussian noise with zero mean and unit variance, and $\overline{\beta_t}$ is controlled by a variance scheduler $\{\beta_t \in (0,1)\}_{t=1}^T$. Diffusion models aim to reverse this diffusion process by sampling random Gaussian noise $x_T$ and gradually denoising the image to generate an image $x_0$. In practice, denoising network $\epsilon_{\theta}$ learns to reverse the diffusion process by predicting the sampled noise $\epsilon$, optimizing the following Mean Square Error:}
\begin{equation}
\mathcal{L}_{mse} = \mathbb{E}_{x_t,c,t,\epsilon \sim \mathcal{N}(0, 1)} \left\|\epsilon-\epsilon_{\theta}\left(x_{t}, c, t\right)\right\|^2_2,
\end{equation}
\hl{where \textit{$x_t$}, \textit{t}, and \textit{c} are the noised image, timestep, and condition, respectively. In this paper, we study with StableDiffusion (SD), a latent diffusion model designed to enhance sampling efficiency while preserving image quality. This method employs an iterative denoising process within a low-dimensional latent space, facilitated by a pre-trained variational autoencoder (VAE). SD runs the reverse process in the latent space to obtain denoised latent feature $\hat{z}_0$, and then decodes $\hat{z}_0$ through the VAE decoder to get the predicted image $\hat{x}_0$.
the objective of diffusion model is simplified as a MSE loss between the sampled noise $\epsilon$ and  by feeding the optimizing the neural network $\epsilon_{\theta}$ by minimizing the  the noised image \textit{$x_t$}, the timestep \textit{t}, and the condition \textit{c} into the:}
\begin{equation}
\mathcal{L}_{mse} = \mathbb{E}_{x_t,c,t,\epsilon \sim \mathcal{N}(0, 1)} \left\|\epsilon-\epsilon_{\theta}\left(x_{t}, c, t\right)\right\|^2_2,
\end{equation}
\hl{where $\epsilon_{\theta}(x_t, c, t)$ is the noise predicted by the diffusion model conditioned on c at timestep \textit{$x_t$}.}

\section{Methodology}\label{sec:method}
Our method introduces an innovative approach, SOEDiff, to significantly enhance the generation of small objects in image synthesis. 
We strategically design a specialized SO-LoRA module (Sec.~\ref{sec:so-lora}) to improve the alignment of image and text within the small area. Complementing this, our approach integrates a cross-scale score distillation loss (Sec.~\ref{sec:cross-scale score distillation}), aimed at enhancing the quality of generated small objects by distilling knowledge from models trained on generic-sized objects. By combining these two techniques, our method achieves a significant improvement in the SOE task.
\begin{figure}[t]
\centering
  \includegraphics[width=\linewidth]{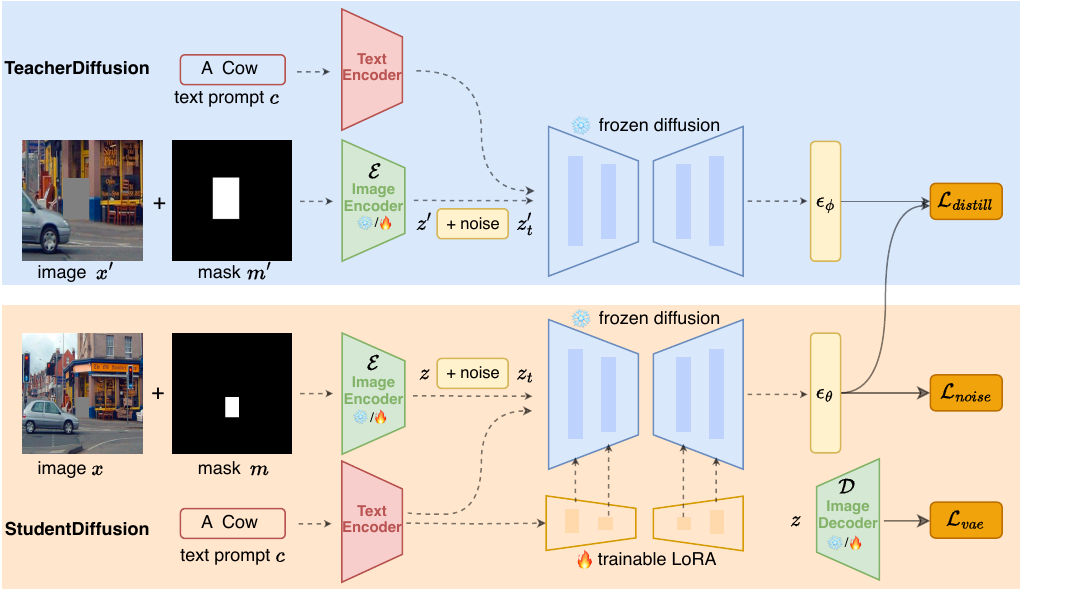}
  \caption{Overview of our proposed SOEDiff. The student diffusion receives image with smaller mask $x$, mask $m$, text prompt $c$ as input and the teacher diffusion receives cropped larger-sized mask image $x'$, mask $m'$, as input to optimize three objectives: (a) denoising loss: the student model aims to predict the noise $\epsilon_\theta$ to match the noise added $\epsilon$ in the diffusion forward process. (b) distillation loss: the student model is trained to generate the same content within the mask region as generated by the teacher model. (c) reconstruction loss: the VAE model is trained to reduce information loss in small target regions.} 
  \Description{}
  \label{fig:trainingpipeline}
\end{figure}

\subsection{Overview}
As depicted in Fig.~\ref{fig:trainingpipeline}, our training pipeline involves multiple networks: a teacher model $\epsilon_{\phi}$, a student model $\epsilon_{\theta}$, and the encoder $\mathcal{E}$ and decoder $\mathcal{D}$ of an autoencoder. We elaborate on each module in our method as follows:

\noindent\textbf{Image Autoencoder:} In line with typical latent diffusion models~\cite{rombach2022high,podell2023sdxl}, the image $x$ (omitting subscript $i$ for clarity) undergoes projection into the latent space with a pre-trained image encoder $\mathcal{E}$ to obtain the latent feature $z$, which could be reconstructed into an image using the image decoder $\mathcal{D}$.

\noindent\textbf{Text Encoder:} Textual description generally serves to control the generated content, leveraging models like CLIP~\cite{radford2021learning} and T5~\cite{colin2020exploring}, pre-trained on large-scale text-image pairs, for text feature extraction. In our SOEDiff approach, we adopt a text encoder from CLIP to encode textual description $c$ into a text feature $\mathcal{E}^{text}(c)$.

\noindent\textbf{SO-LoRA:} Our method incorporates two diffusion models initialized with pre-trained SD/SDXL. During the training stage, the teacher diffusion model $\epsilon_{\phi}$ remains frozen, while we integrate the adapter SO-LoRA with the student diffusion model $\epsilon_{\theta}$, in which only parameters of SO-LoRA are updated for fast adaptation.

\noindent\textbf{Optimization:} We introduce three objective losses, denoted as $\mathcal{L}_{denoise}$, $\mathcal{L}_{distill}$, and $\mathcal{L}_{vae}$, to optimize the image autoencoder and the SO-LoRA module. For $\mathcal{L}_{denoise}$, we apply the diffusion process on the latent feature $z$ as described in Eq.~\ref{eq:diffusion process} to obtain the noised latent $z_t$, which is then fed into the student model to predict the noise $\epsilon$ with $\epsilon_\theta(z_t, c, t, m)$. 

For $\mathcal{L}_{distill}$, we begin by center-cropping the image around the target object to obtain the re-scaled image $x'$ and mask $m'$, which are then input into the teacher diffusion model to acquire the re-scaled predicted noise $\epsilon_{\phi}(z'_t, c, t, m')$, where $z'_t$ is latent feature extracted by the image encoder. We enforce the prediction from the student model to be close to the prediction from the teacher's by minimizing $\mathcal{L}_{distill}$. 
Finally, since the VAE model is critical for generating high-fidelity small objects, we further fine-tune the encoder and decoder to enhance detail preserving with a reconstruction loss $\mathcal{L}_{vae}(x, \mathcal{D}(\mathcal{E}(x)))$, where we use the Huber loss for robust regression. Thus, we optimize the network with a total loss:
\begin{equation}
\begin{aligned}\mathcal{L}_{total} = & \mathcal{L}_{denoise}(\epsilon, \epsilon_\theta(z_t, c, t, m)) + \alpha \mathcal{L}_{distill}(\epsilon_\theta(z_t, c, t, m), \epsilon_{\phi}(z'_t, c, t, m')) \\&+ \lambda \mathcal{L}_{vae}(x, \mathcal{D}(\mathcal{E}(x))).    \end{aligned}
\label{eq:total loss}
\end{equation}
We will subsequently elaborate on these objective losses in the following sections.

\subsection{SO-LoRA}\label{sec:so-lora}
\hl{SO-LoRA is designed to efficiently adapt pre-trained diffusion models to learn scale-invariant representations for the SOE task. Inspired by}~\cite{qin2021exploring}\hl{, the SOE task, being a downstream task of image editing, can be re-parameterized by optimizing only a few free parameters within a unified low-dimensional intrinsic task space. Consequently, we fine-tune only a subset of the parameters in the pre-trained model to balance performance and efficiency.}

In designing our SO-LoRA module, we have adhered to the principle established in LCM~\cite{luo2023latent} and ControlNet~\cite{zhang2023adding} designs. The structure of our SO-LoRA is the same as the backbone network. For a specific parameter weight $W_0\in {\mathbb{R}^{j \times k}}$, where $j$ is the input dimension and $k$ is the output dimension, the weight update $\Delta W$ is decomposed into two low-rank matrices $\Delta W = BA$, where $B\in {\mathbb{R}^{j \times r}}, A \in {\mathbb{R}^{r \times k}}$ and $r \ll min(j, k)$. During initialization, the entries of $B$ are all set to zero to ensure $\Delta W=0$ at the beginning of training, and $A$ is initialized with Gaussian distribution. By freezing $W_0$ and optimizing only the matrices $A$ and $B$, LoRA achieves significant reductions in trainable parameters. During inference, $\Delta W$ can be merged into $W_0$ with no additional computation/memory overhead using a scaling factor $\alpha$:
\begin{equation}
   W=W_0+\alpha \Delta W = W_0 + \alpha BA,
\end{equation}
during the training phase, only the LoRA modules in the student model $\epsilon_{\theta}$ are trainable.

Given an input image $x$ and the corresponding mask $m$, we encode the image $x$ with the encoder $\mathcal{E}$ to get latent feature $z$ and progressively add noise to $z$ to yield a noisy latent $z_t$, where $t$ represents the timestep. 

To refine SO-LoRA for the task of generating small objects, denoising loss is calculated within the masked area. This focus ensures that SO-LoRA's training and subsequent operations are concentrated on the small target regions, so that it can precisely enhance alignment between the target region and the corresponding textual description. The denoising loss is formulated as follows:
\begin{equation}
    \mathcal{L}_{denoise} = \mathbb{E}_{z_{t}, t, c, m, \epsilon \sim \mathcal{N}(0, 1)}
        \mathcal{L} (\epsilon \odot m, \epsilon _{\theta}(z_{t}, t,c , m) \odot m ),
\label{eq:denoising loss}
\end{equation}
where $\mathcal{L}$ is the Huber loss.

\subsection{Cross-scale Score Distillation}\label{sec:cross-scale score distillation}
To enhance the quality of generated small object images, we introduce an additional teacher model with weight $\phi$ for cross-scale distillation, leveraging high-resolution predictions as guidance. 
As previously mentioned, the student model takes the noised image feature $z_t$, mask $m$, and corresponding text prompt $c$ as inputs, then predicts noise $\epsilon_{\theta}(z_{t}, t, c, m)$.
For the teacher model, we first crop a region with a size of $s$ centered around the masked area of the original image $x$ and resize it to a new image $x'$ with the same size as $x$. Then, we feed $x'$ into the image encoder $\mathcal{E}$ to obtain $z'$. Similar to the student model, we obtain the predicted noise $\epsilon_{\phi}(z'_{t}, t, c, m')$. In this enlarged masked area, we assume that more details are preserved and the prediction of the teacher model is more reliable. Therefore, we consider the reverse diffusion process as an augmented Probability Flow Ordinary Differential Equation (PF-ODE) following LCM~\cite{luo2023lcm}, and then revert the predicted noise $\epsilon_{\theta}(z_{t}, t, c, m)$ (\textit{resp.} $\epsilon_{\phi}(z'_{t}, t, c, m')$) to image $\hat{z}_0$ (\textit{resp.} $\hat{z}'_0$) as follows:
\begin{equation}
\label{predictx0}
    \hat{z}_0 = \frac {z_t-\sigma_t \epsilon_\theta(z_t, t, c, m)}{\alpha_t},
\end{equation}
where $\sigma_t$ and $\alpha_t$ are calculated from the predefined variance scheduler $\beta$. Next, features corresponding to the masked area are cropped from predictions $\hat{z}_0$ and $\hat{z}'_0$. Then we resize these two features and align them with the following distillation loss:
\begin{equation}
    \mathcal{L}_{distill} = \mathcal{L}(r(\hat{z}_0 \odot m), r(\hat{z}'_0 \odot m')),
\label{eq:distillation loss}
\end{equation}
where $r(\cdot)$ denotes the operation to resize the cropped latent feature to the same size, and $\mathcal{L}$ is the Huber loss. We summarize the algorithm of SOEDiff in Alg.~\ref{alg:SOEDiff}. 

\begin{algorithm}[t]
    \caption{Training procedure of SOEDiff}
    \label{alg:SOEDiff} 
    \SetAlgoNoLine
        \textbf{Input:} teacher model parameter $\phi$, student model parameter $\theta$, autoencoder $\mathcal{E}(\cdot)$, $\mathcal{D}(\cdot)$, learning rate $\eta$, noise schedule $\alpha_t$, $\sigma_t$, timestep $t$, distance measurement $\mathcal{L}(\cdot, \cdot)$ \\
        
         \vspace{1mm} \hrule \vspace{1mm}
         \textbf{repeat} \\
        \qquad Sample image $x$, mask $m$, prompt $c$ from dataset \\
        \qquad Crop and resize $x$ to obtain $x'$ \\
        \qquad Crop and resize $m$ to obtain $m'$ \\
        \qquad Encode images to latent space $z \gets \mathcal{E}(x)$, $ z' \gets \mathcal{E}(x')$  \\
        \qquad Calculate the reconstruction loss $\mathcal{L}_{vae} = \mathcal{L}(x\odot m, \mathcal{D}(\mathcal{E}(x)) \odot m)$ \\
        
        \qquad Add noise to the latent features $z_{t} \sim \mathcal{N}(\alpha_tz;\sigma^2_t\mathbf{I})$, $z'_{t} \sim \mathcal{N}(\alpha_tz';\sigma^2_t\mathbf{I})$ \\
    
        \qquad Calculate the denoising loss $\mathcal{L}_{denoise}$ as presented in Eq.~\ref{eq:denoising loss} \\
        \qquad Revert noise to the image $\hat{z}_0 \gets f_\theta(z_t, t, c, m)$, $\hat{z}'_0 \gets f_\phi(z'_t, t, c, m')$ \\

        \qquad Calculate the distillation loss $\mathcal{L}_{distill}$ as presented in Eq.~\ref{eq:distillation loss}\\
    
        \qquad $\theta \gets \theta - \eta \nabla_\theta    (\mathcal{L}_{denoise} + \mathcal{L}_{distill} + \mathcal{L}_{vae}) $ \\
    
        \textbf{until} convergence\\

\end{algorithm}

\section{Experiments}\label{sec:experiments}
\subsection{Datasets and Evaluation Metrics}
\noindent\textbf{Training Data.} 
To build our training dataset, we curated 300K image data from the vast \textit{OpenImage} dataset~\cite{benenson2019large}, imposing rigorous standards to guarantee the data's suitability for our small object editing task. We selected objects meeting specific criteria: they had to be unobscured, fully contained within the image boundaries, and represent tangible real-world entities. Furthermore, these objects had to exist individually rather than as part of a larger group, and they must not include human body parts like eyes or mouths. Crucially, the area occupied by these objects had to be less than $(1/8)^2$ of the total image area.

\noindent\textbf{Validation Data.} For our validation data, we carefully curated subsets from the MSCOCO-val~\cite{lin2014microsoft} and OpenImage-val datasets, specifically focusing on data where object sizes range between $(1/8)^2$ and $(1/6)^2$ of the image area. This selection process adheres to the same criteria used for curating the training dataset. The selected test data ensures that objects are unobstructed by other elements, resulting in two customized test datasets: MSCOCO-small-val and OpenImage-small-val.
Moreover, we cropped the images to extract regions within the masked areas and employed the BLIP-VQA~\cite{li2022blip} model to inquire about the primary color of the object in each area. This method efficiently identifies the predominant colors of objects within the masks.
The OpenImage-small-val dataset encompasses over 250 common categories, totaling approximately 2,000 images. In comparison, the MSCOCO-small-val dataset comprises images from 80 distinct categories, also amounting to about 2,000 images. Furthermore, we incorporated the complete MSCOCO-val and OpenImage-val datasets as additional validation datasets. For reproducibility, we will make the training and validation data public.

\noindent\textbf{Size Threshold Analysis.} Upon analysis, we identified the thresholds of 1/6 and 1/8 as critical for training and evaluation on small object editing. Our objective is to accurately generate small objects within this threshold. When the mask's side length occupies 1/5 of the image size, the corresponding $8\times8$ cross-attention map should ideally have feature side lengths of at least $round(\frac{8}{5})=2$. Similarly, if the mask's side length is 1/6 of the original image, the corresponding feature size decreases to $1\times1$. In scenarios where the side length is smaller than 1/8, it is even plausible that no corresponding features exist. The model's training data comprises challenging examples aimed at enhancing its learning and performance. Furthermore, this approach ensures a more equitable validation process, given that the model has not been exposed to objects sized between 1/8 and 1/6 during the training phase.

\noindent\textbf{Evaluation Metrics.}
We utilize the Fr'echet Inception Distance (FID) to measure image quality and CLIP-Score to assess text-to-image alignment, following the evaluation protocol used in DDPM~\cite{ho2020denoising}. Higher CLIP-Score or lower FID values indicate superior performance. Importantly, we compute these metrics specifically on the masked areas, rather than employing the entire image for evaluation. This evaluation protocol is motivated by the fact that small objects occupy a relatively minor proportion of the overall image. Calculating metrics using the entire image could disproportionately bias the evaluation towards large, irrelevant areas, thus distorting the assessment of the model's performance. 

\begin{table}[h]
  \caption{\hl{Quantitative results on OpenImage-small-val and MSCOCO-small-val. Our proposed method SOEDiff could be combined with the existing approaches, \ie SD1.5, and SDXL, we denote these methods as SOEDiff, and SOEDiff-XL. The experimental results indicate that our proposed method significantly surpass the baselines, and achieves state-of-the-art performance on these two benchmarks. The best results are in~\textbf{bold}.}}
  \resizebox{\columnwidth}{!}{
    \begin{tabular}{ccccccc}
    \toprule
    \multirow{2}{*}{Prompt Style} & \multirow{2}{*}{Base Model} & \multirow{2}{*}{Methods} & \multicolumn{2}{c}{OpenImage-small-val}  & \multicolumn{2}{c}{MSCOCO-small-val} \\\cline{4-7}
                & & & CLIP-Score $\uparrow$    & FID $\downarrow$  & CLIP-Score $\uparrow$   & FID $\downarrow$    \\
    \hline
    \multirow{14}{*}{Label Only}
              & \multirow{9}{*}{SD1.5 Inpainting}
              & Blended DM  & 23.66 & 43.20   & 22.83   & 38.78     \\
              & &SD-I & 23.79 & 34.65   & 22.55   & 30.78     \\
              & & P2P & 23.88 & 34.76 & 22.98 & 35.11 \\
              & & LGG(SD-I) & 24.05 & 34.32 & 23.39 & 34.67 \\
              & &Inpaint-Anything  & 24.09 & 33.52   & 22.81   & 29.38     \\
              & & LCM (4-steps) & 24.37 & 34.65 & 23.14 & 30.56 \\
              & & LCM (20-steps) & 24.62 & 33.86 & 23.53 & 30.14 \\
              & & SD-I (Teacher)  & 24.51 & 32.84   & 23.42   & 30.21     \\\cline{3-7}
              & & \textbf{SOEDiff-SD (Ours)} & \textbf{24.78} & \textbf{31.78}   & \textbf{23.89}   & \textbf{29.40}    \\\cline{2-7}
              & \multirow{2}{*}{DreamShaper Inpainting} & DreamShaper & 24.58 & 33.77 & 	
              23.15 & 30.14\\
              & & HD-Painter & 24.39 & 32.74   & 23.25   & 29.97   \\\cline{2-7}
              & \multirow{3}{*}{SDXL Inpainting} & LGG(SDXL-I) & 24.46 & 33.67 & 23.96 & 34.02 \\
              & &SDXL-I  & 25.56 & 34.26   & 24.50   & 31.26     \\\cline{3-7}
              & &\textbf{SOEDiff-SDXL (Ours)} & \textbf{26.17}  & \textbf{31.33}   & \textbf{25.32}   & \textbf{29.25}\\
    \hline
    \multirow{8}{*}{\makecell[c]{Color + Label}}
              & \multirow{7}{*}{SD1.5 Inpainting} 
                &Blended DM        & 23.85 & 43.26   & 22.94   & 38.01     \\
              & &SD-I              & 24.13 & 34.73   & 22.66   & 30.65     \\
              & &Inpaint-Anything  & 24.50 & 33.69   & 23.02   & 28.92     \\
              & &LCM (4steps) & 	24.97 & 34.29 & 23.45 & 30.30 \\
              & &LCM (20steps) & 25.15 & 34.10 & 23.74 & 29.26 \\
              & &SD-I (Teacher)   & 25.11 & 32.45   & 24.03   & 28.49     \\
              & &\textbf{SOEDiff-SD (Ours)}           & 25.57 & 31.21   & 24.38   & \textbf{27.62}     \\\cline{2-7}
              & \multirow{2}{*}{DreamShaper Inpainting} & DreamShaper & - & - & 23.61 & 29.42 \\
              & &HD-Painter         & 25.19 & 32.35   & 23.98   & 29.01     \\\cline{2-7}
              & \multirow{2}{*}{SDXL Inpainting} &SDXL-I & 25.81 & 34.87   & 24.46   & 31.23     \\
              & &\textbf{SOEDiff-SDXL (Ours)}        & \textbf{26.83} & \textbf{31.11}   & \textbf{25.52}   & 29.07     \\   
    \bottomrule
    \end{tabular}
  }
  \label{tab:mainexp}
  \end{table}

\subsection{Text-based Small Object Editing}\label{sec:exp:text-based_small_object_editing}
\noindent\textbf{Implementation Details}:
For our experiments, we employed Stable-Diffusion-v1-5-Inpainting (SD-I) and Stable-Diffusion-xl-1.0-inpainting (SDXL-I) as the base models, along with the DDIM-Solver for fast inference. We maintained a batch size of 8 and conducted training for 100K steps. It's worth noting that fine-tuning the VAE concurrently significantly increases computation cost, necessitating a reduction in batch size and prolonging the training duration. Regarding loss weights, we set $\alpha$ to 0.01 and $\lambda$ to 1 to ensure that the initial losses of all components are aligned at the same order of magnitude. Additionally, the crop size $s$ was set to 256 in our experiments.

\noindent\textbf{Quantitative Result}:
We compared models trained on both backbones, \hl{SD-I, DreamShaper, and SDXL-I, alongside the mask-conditioned object generation model,} BlendedDM~\cite{avrahami2022blended} (based on SD-1.5), \hl{P2P}~\cite{hertz2022prompt}, \hl{LGG}~\cite{pan2024towards}, Inpaint-Anything~\cite{yu2023inpaint} (based on SD-2-Inpainting), DreamShaper Inpainting~\cite{dreamshaper}, \hl{LCM}~\cite{luo2023lcm}, and HD-Painter~\cite{manukyan2023hd}. \hl{P2P and LGG are methods based on attention modification, DreamShaper is a widely used model in the community, and LCM is a recent work known for its competitive performance in few-step editing. We applied SOEDiff on SD-I and SDXL-I, denoted as SOEDiff-SD and SOEDiff-SDXL, respectively.} Additionally, we evaluated them on our small-sized object validation datasets, OpenImage-small-val and MSCOCO-small-val, using two types of prompt templates: Label Only (\eg, ``a dog'') and Color+Label (\eg, ``a brown dog''). The results are shown in Tab.~\ref{tab:mainexp}. Our method demonstrated a significant improvement compared to the baseline models on these two datasets, affirming the efficacy and superiority of our approach for small object generation. 
\hl{The lightweight SOEDiff (SOEDiff-SD in the table) is developed based on the SD1.5 Inpainting model. The improvement of SOEDiff-SD over the existing work developed on SD1.5, such as Inpaint-Anything and LCM, is significant. Specifically, on the OpenImage-small-val, SOEDiff-SD outperforms Inpaint-Anything by 0.69 in CLIP-Score and 1.74 in FID, and outperforms LCM by 0.41 in CLIP-Score and 1.09 in FID. On the MSCOCO-small-val, SOEDiff-SD outperforms Inpaint-Anything by 1.08 in CLIP-Score and 1.18 in FID, and outperforms LCM (4-steps) by 0.75 in CLIP-Score and 1.16 in FID. Moreover, SOEDiff-SD even outperforms the teacher model used for distillation (SD-I (Teacher) in the table) by 0.27 in CLIP-Score and 1.06 in FID on the OpenImage-small-val. These results demonstrate the effectiveness of the proposed SOEDiff on the SOE task.}
Some questions may arise regarding instances where the FID scores of the SDXL model appear lower in certain comparative scenarios such as the MSCOCO-small-val dataset. This phenomenon has also been observed in SDXL~\cite{podell2023sdxl}. It's essential to note that higher FID scores do not necessarily correlate with worse visual outcomes. Furthermore, the $512\times512$ images generated by SDXL, which can lead to poorer outcomes, may also contribute to these discrepancies. 
In addition to comparisons with state-of-the-art methods, we also evaluate the performance of the teacher model in generating small-sized objects, similar to the approach mentioned in the article where the teacher processes images by cropping the original image. The new image obtained has a mask area with sides at least twice the length of the original, transforming the small object generation issue into a somewhat simpler problem. This is referred to as the ``Teacher'' in the table.

In addition to testing on the small-sized object validation dataset, we conducted evaluations on the full-size OpenImage-val and MSCOCO-val datasets. This allowed us to demonstrate that our method is not only effective for small target objects but also beneficial for enhancing generic-sized object inpainting tasks. The results are presented in Tab.~\ref{tab:unfilterresult}. For the ``Label Only'' prompt, SOEDiff outperforms SD-I on both datasets. It achieves a CLIP-Score of 25.52 compared to 25.08 by SD-I on OpenImage-val, and 23.68 versus 23.09 on MSCOCO-val. Similarly, SOEDiff shows superior FID scores, recording 6.80 on OpenImage-val and 6.30 on MSCOCO-val, improvements over the 7.81 and 6.88 recorded by SD-I, respectively.
The ``Color + Label'' prompt, which presumably includes additional color information alongside the object labels, shows an overall improvement for both methods across both metrics and datasets compared to the ``Label Only'' prompt alone. SOEDiff again leads with the highest scores: a CLIP-Score of 26.89 and FID of 6.69 on OpenImage-val, and 24.07 and 6.22 on MSCOCO-val. These results underline SOEDiff's effectiveness in handling more complex prompts.

\begin{table}[t]
  \caption{\hl{Quantitative results on full-size test datasets OpenImage-val and MSCOCO-val. The best results are in \textbf{bold}.}}
  \begin{tabular}{cccccc}
  \toprule
  \multirow{2}{*} {Prompt Style} & \multirow{2}* {Methods} & \multicolumn{2}{c}{OpenImage-val}  & \multicolumn{2}{c}{MSCOCO-val} \\\cline{3-6} 
              &  & CLIP-Score $\uparrow$      & FID $\downarrow$  & CLIP-Score $\uparrow$   & FID $\downarrow$    \\ 
          \hline
  \multirow{3}{*}{Label Only}          
              &SD-I          & 25.08 & 7.81   & 23.09    & 6.88    \\
              &SD-I w/ SO-LoRA & 25.40 & 7.23 & 23.54 & 6.44 \\
              &SOEDiff     & \textbf{25.52} & \textbf{6.80}   & \textbf{23.68}   & \textbf{6.30}     \\
  \hline
  \multirow{3}{*}{\makecell[c]{Color + Label}}    
              &SD-I        & 26.21 & 7.72   & 23.23   &  6.65    \\
              &SD-I w/ SO-LoRA & 26.67 & 6.94 & 23.72 & 6.28 \\
              &SOEDiff     & \textbf{26.89} & \textbf{6.69}   & \textbf{24.07}   & \textbf{6.22}     \\
  \bottomrule
  \end{tabular}
  \label{tab:unfilterresult}
  \end{table}

\begin{table}[t]
\caption{Effectiveness of the components used in SOEDiff. We employ SD-I as the base model for comparison. SD-I denotes StableDiffusion-v1.5-inpainting, SO-LoRA is the proposed LoRA module for small object editing, CSD corresponds to cross-scale distillation, and VT is VAE tuning for enhancing the object details.}
\begin{tabular}{cccccccc}
\toprule
\multicolumn{4}{c}{ Component}                 & \multicolumn{2}{c}{OpenImage-small-val}   & \multicolumn{2}{c}{MSCOCO-small-val} \\ 
\hline
SD-I & SO-LoRA & CSD & VT   & CLIP-Score $\uparrow$      & FID $\downarrow$  & CLIP-Score $\uparrow$   & FID $\downarrow$    \\ 
\hline
\checkmark & & &                                    & 24.10    & 34.73  & 22.66 & 30.65   \\
\checkmark &\checkmark & &                          & 24.93       & 33.78  & 23.53 & 29.60   \\
\checkmark &\checkmark & & \checkmark               & 25.16       & 33.98  & 23.67 & 29.24   \\
\checkmark &\checkmark & \checkmark&                & 25.40       & 32.41  & 24.14 & 28.40   \\
\checkmark &\checkmark &\checkmark &\checkmark      & 25.58       & 31.21  & 24.38 & 27.62   \\
\bottomrule
\end{tabular}
\label{tab:ablation}
\end{table}

\begin{figure}[t]
\centering
  \includegraphics[width=0.99\linewidth]{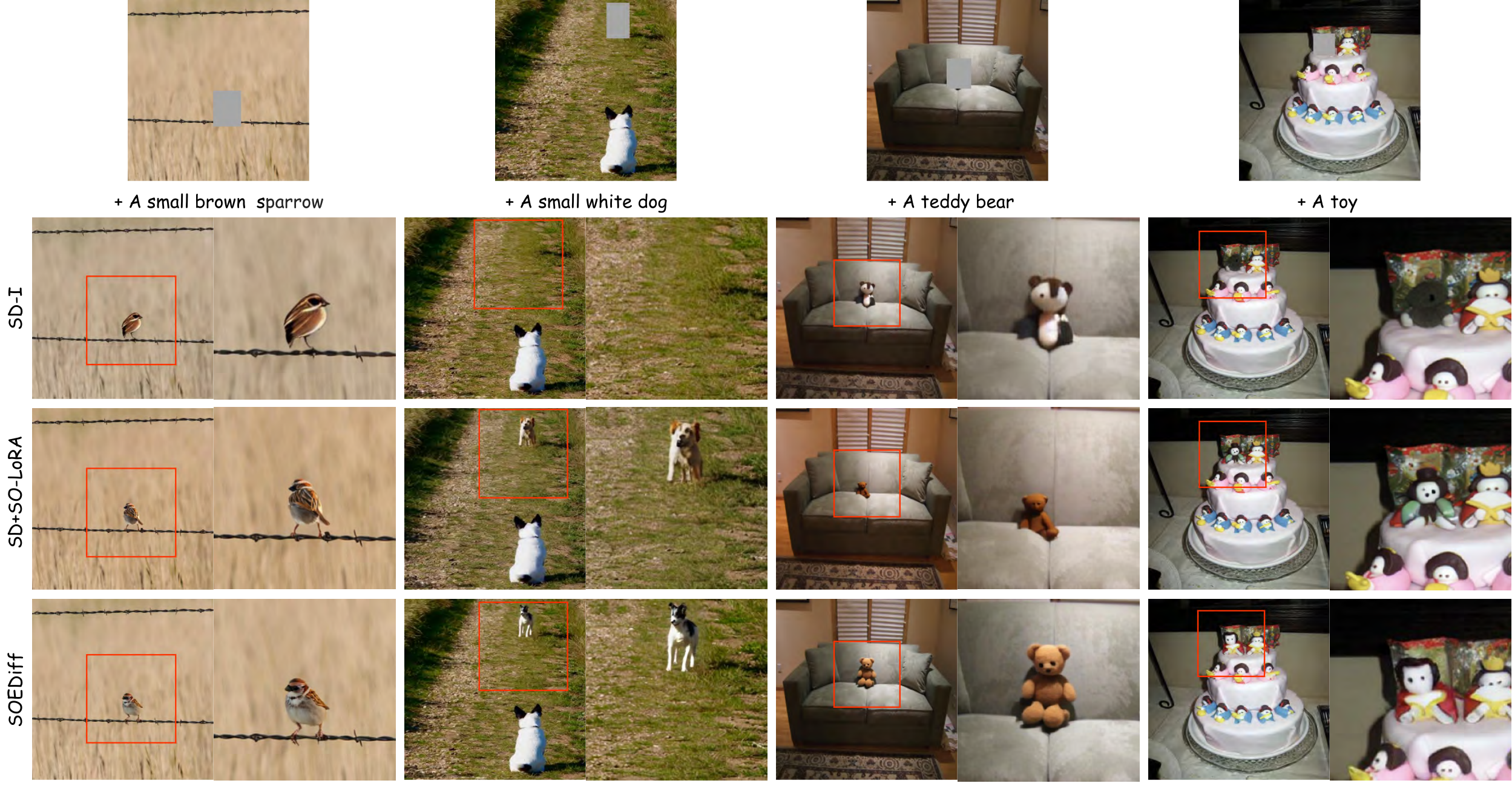}
  \caption{Qualitative comparison of different components: the top row displays masked images and their corresponding text prompts. The second row shows edited images generated by SD-I. In the third and fourth rows, results from SD+SO-LoRA and SOEDiff are presented.}
  \Description{}
  \label{fig:ablationimage}
\end{figure}

\noindent\textbf{Effectiveness of Modules}:
As reported in Tab.~\ref{tab:ablation}, we validate the effectiveness of each module by testing different combinations of them with color+label as text prompt, focusing on baseline model SD-I, SO-LoRA, Cross-scale Distillation (CSD), and VAE Tuning (VT) across two datasets: OpenImage-small-val and MSCOCO-small-val. Initial configurations with SD-I alone yielded CLIP-Scores of 24.10 and FID of 34.73 on OpenImage-small-val, with corresponding scores of 22.66 and 30.65 on MSCOCO-small-val. The addition of SO-LoRA improved the CLIP-Scores to 24.93 and reduced FID to 33.78 on OpenImage, while MSCOCO scores improved to 23.53 and 29.60, respectively. Incorporating VT alongside these components further enhanced the scores marginally to 25.16 and 33.98 in FID on OpenImage-small-val, and slightly better performance on MSCOCO-small-val. The integration of CSD, with or without VT, yielded the most significant enhancements: with a combined setup, the OpenImage-small-val results showed a peak CLIP-Score of 25.58 and the lowest FID at 31.21. On MSCOCO-small-val, the model achieved its best performance with a CLIP-Score of 24.38 and an FID of 27.62. These numerical improvements highlight the effectiveness of each component, illustrating their synergistic impact in refining both the alignment and quality of the generated images, thus validating our modular enhancement strategy in the development of advanced text-to-image generation technologies. Except for the quantitative results, we present visual comparisons in Fig.~\ref{fig:ablationimage}, we find that the incorporation of SO-LoRA notably improves text-to-image alignment and mitigates the aforementioned issues (\ie, Object Missing, Text-Image Mismatch, and Distortion).

\noindent\textbf{Effectiveness of SO-LoRA}: 
\hl{To validate the effectiveness of SO-LoRA, we conducted an ablation study on different blocks for fine-tuning. As shown in Table}~\ref{tab:exp:ablation_solora}, \hl{the results demonstrate that fine-tuning all blocks achieves the best performance on the SOE task. Specifically, we experimented with various combinations of mid, down, and up blocks. The results indicate that incorporating all blocks (Mid, Down.2+Up.2, Down.1+Up.1, and Down.0+Up.3, the U-Net is consisted of four DownBlocks, one MidBlock, and four UpBlocks, here we denote the ``DownBlock'' as ``Down'', ``MidBlock'' as ``Mid'', ``UpBlock'' as ``Up'') consistently yields the highest CLIP-Scores and the lowest FID values on both OpenImage-small-val and MSCOCO-small-val datasets. This comprehensive fine-tuning approach significantly enhances the model's ability to generate high-quality, text-aligned small objects, confirming the importance of each block in the SO-LoRA module.}

\begin{table}[h]
  \centering
  \caption{\hl{Ablation study for the SO-LoRA module. The results are reported on the OpenImage-small-val and MSCOCO-small-val. The best results are in~\textbf{bold}.}}
  \resizebox{\columnwidth}{!}{
    \begin{tabular}{ccccccccc}
    \toprule
    \multirow{2}{*}{Prompt Style} & \multirow{2}{*}{Mid} &\multirow{2}{*}{Down.2+Up.2} &\multirow{2}{*}{Down.1+Up.1} &\multirow{2}{*}{Down.0+Up.3} &\multicolumn{2}{c}{OpenImage-small-val}  &\multicolumn{2}{c}{MSCOCO-small-val} \\\cline{6-9}
    &  &    &    &   & CLIP-Score$\uparrow$ & FID$\downarrow$  & CLIP-Score$\uparrow$ & FID$\downarrow$ \\\hline
    \multirow{8}{*}{Label Only} & &  &  &  & 23.79  & 34.65  & 22.55  & 30.78 \\
    &\checkmark &  &  &  & 24.27 & 33.50 & 23.01 & 30.17 \\ 
    &\checkmark & \checkmark &  &  & 24.35 & 33.30 & 23.19 & 29.88 \\
    &\checkmark & \checkmark & \checkmark & & 24.62 & 31.92 & 23.69 & 30.03 \\
    & &  &  & \checkmark & 24.17 & 33.54 & 23.02 & 30.09 \\
    & &  & \checkmark & \checkmark & 24.37 & 32.19 & 23.26 & 29.72 \\
    & & \checkmark & \checkmark & \checkmark & 24.70 & 31.87 & 23.62 & 29.51 \\
    & \checkmark & \checkmark & \checkmark & \checkmark & \textbf{24.78} & \textbf{31.78} & \textbf{23.89} & \textbf{29.40} \\\hline
    \multirow{8}{*}{Color + Label} & &  &  &  & 24.13 & 34.73  & 22.66 & 30.65 \\
    &\checkmark &  &  &  & 24.59 & 33.33 & 23.09 & 29.10 \\
    &\checkmark & \checkmark &  &  & 24.79 & 33.19 & 23.50 & 28.44 \\
    &\checkmark & \checkmark & \checkmark & & 25.49 & 31.57 & 24.14 & 27.72 \\
    & &  &  & \checkmark & 24.69 & 33.48 & 23.12 & 29.19 \\
    & &  & \checkmark & \checkmark & 24.95 & 32.39 & 23.59 & 28.36 \\
    & & \checkmark & \checkmark & \checkmark & 25.39 & 31.46 & 24.02 & 27.75 \\
    & \checkmark & \checkmark & \checkmark & \checkmark &\textbf{25.57} &\textbf{31.21} & \textbf{24.38} &\textbf{27.62} \\ 
    \bottomrule
    \end{tabular}
  }
  \label{tab:exp:ablation_solora}
\end{table}

\begin{table}[t]
\caption{CLIP-Score computed with different loss weight $\lambda$ and different crop size $s$ on OpenImage-small-val and MSCOCO-small-val.}
\begin{tabular}{ccccc}
\toprule
$\lambda$  & 0.1 & 0.01 & 0.005 & 0.001   \\\hline
OpenImage-small-val & 25.08 & 25.40 & 25.32 & 25.21 \\
MSCOCO-small-val  & 23.83 &24.38 & 24.28 & 24.16 \\
\midrule
$s$  & 128  & 128$\sim$256 & 256 &  256$\sim$512  \\\hline
OpenImage-small-val& 25.25  & 25.33  &25.40 & 25.36 \\
MSCOCO-small-val &  24.12     &24.30       &24.38 &24.28\\
\bottomrule
\end{tabular}
\label{tab:sablation}
\end{table}

\begin{figure}[t]
\centering
  \includegraphics[width=\linewidth]{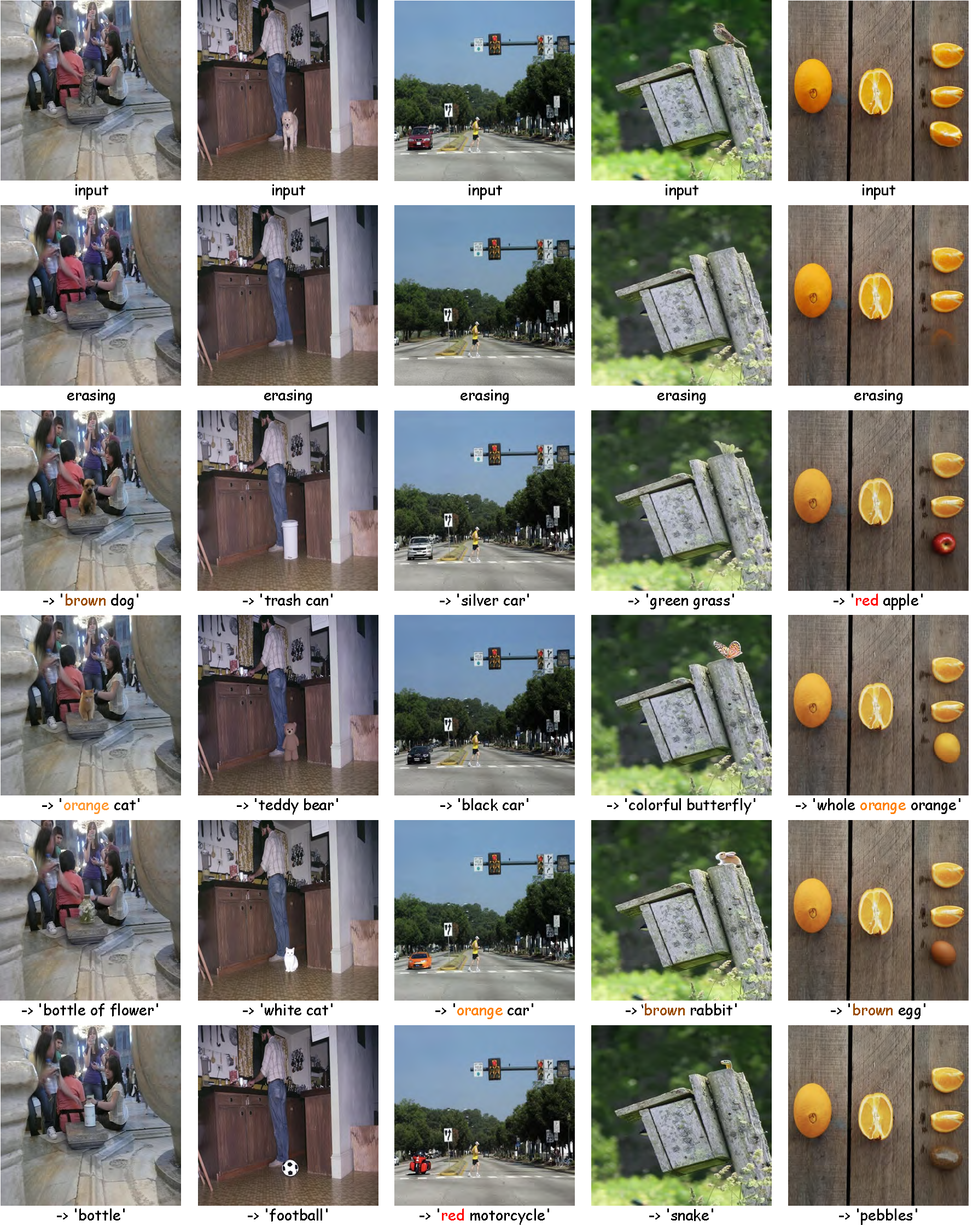}
  \caption{Extended application with our proposed SOEDiff. The first row shows the original images, the second row displays the results of object erasing, and rows three to six depict the results of object replacement.}
  \Description{}
  \label{fig:moreextend}
\end{figure}

\noindent\textbf{Analysis of Loss Weight $\alpha, \lambda$ and Crop Scale $s$}:
\hl{In our study, we meticulously examined the impact of three critical hyper-parameters: loss weights $\alpha$ and $\lambda$, and crop size $s$, as shown in Tab.}~\ref{tab:exp:ablation_huber_loss} and Tab.~\ref{tab:sablation}. \hl{These parameters were evaluated for their influence on the performance of our image generation model, particularly when using cropped images of small objects as inputs into the teacher model. The experimental results reveal that variations in crop size $s$ had a limited effect on CLIP-Scores. However, adjustments to the loss weights $\alpha$ and $\lambda$ were pivotal. }

\hl{We compared the performance of Huber Loss and MSE Loss for distillation and identified the optimal $\alpha$. The findings indicate that Huber Loss outperforms MSE Loss, with the best performance achieved at $\alpha=1$ for $\mathcal{L}_{distill}$. When $\lambda$ was set to 0.01, the model reached peak performance with a CLIP-Score of 25.40 on OpenImage-small-val and 24.38 on MSCOCO-small-val. Conversely, increasing $\lambda$ to 0.1 or decreasing it to 0.001 led to a significant drop in performance, highlighting the sensitivity of this parameter. These results underscore the importance of fine-tuning $\lambda$ to achieve optimal model performance, as inappropriate values can impede effective optimization and degrade overall results.}

\begin{table}[h]
  \centering
  \caption{\hl{Ablation study for the distillation loss $\mathcal{L}_{distill}$. The results are reported on the OpenImage-small-val and MSCOCO-small-val. The best results are in~\textbf{bold}.}}
    \begin{tabular}{ccccccc}
      \toprule
      \multirow{2}{*}{Prompt Style} &\multirow{2}{*}{Loss Type} &\multirow{2}{*}{$\alpha$} &\multicolumn{2}{c}{OpenImage-small-val} &\multicolumn{2}{c}{MSCOCO-small-val} \\\cline{4-7}
                                    &      &     & CLIP-Score$\uparrow$ & FID$\downarrow$ & CLIP-Score$\uparrow$ & FID$\downarrow$ \\\hline
      \multirow{5}{*}{Label Only} & \multirow{3}{*}{Huber Loss} & 1 & \textbf{24.78}       & \textbf{31.78}  & \textbf{23.89}       &\textbf{29.40} \\
                         &     & 0.5                        & 24.72                &31.82            & 23.53                &29.58 \\
                         &     & 0.1                        & 24.53                &32.01            & 23.40                &29.19 \\\cline{2-7}
                         & \multirow{2}{*}{MSE Loss} & 1 & 24.51 &32.12 & 23.45 &29.70 \\
                         &        & 0.5 & 24.46 &32.24 & 23.27 &29.96 \\\hline
      \multirow{5}{*}{Color + Label} & \multirow{3}{*}{Huber Loss} & 1 & \textbf{25.57}       & \textbf{31.21}  & \textbf{24.38}       &\textbf{27.62} \\
                         &     & 0.5                        & 25.47              &32.05              & 23.96                &28.11 \\
                         &     & 0.1                        & 25.15            &32.20              & 23.82                &28.34 \\\cline{2-7}
                         & \multirow{2}{*}{MSE Loss} & 1 & 25.17 &32.02 & 23.77 &28.10 \\
                         &        & 0.5 & 25.01 &32.16 & 23.45 &28.16 \\
      \bottomrule
    \end{tabular}
    \label{tab:exp:ablation_huber_loss}
\end{table}

\noindent\textbf{Extended Applications}: Beyond its core functionality, SOEDiff excels in object removal and replacement, demonstrating its versatility across various image editing tasks. This feature is crucial for applications such as digital forensics, content moderation, and creative media production.
\begin{itemize}
\item\textbf{Object Removal}: SOEDiff can efficiently erase unwanted objects from images when provided with a targeted textual prompt, streamlining content editing for clearer visual communication.
\item\textbf{Object Replacement}: The model supports seamless replacement of objects as directed by user input, facilitating creative alterations like changing elements within a scene to suit different thematic or aesthetic requirements.
\end{itemize}
These capabilities are exemplified in Fig.~\ref{fig:moreextend}, which showcases the model’s proficiency in maintaining the natural look and coherence of the background post-editing. This flexibility not only enhances the model’s applicability to diverse editing needs but also simplifies complex tasks, making advanced image manipulation more accessible to a broader audience.

\begin{figure}[t]
  \begin{center}
  \includegraphics[width=0.95\linewidth]{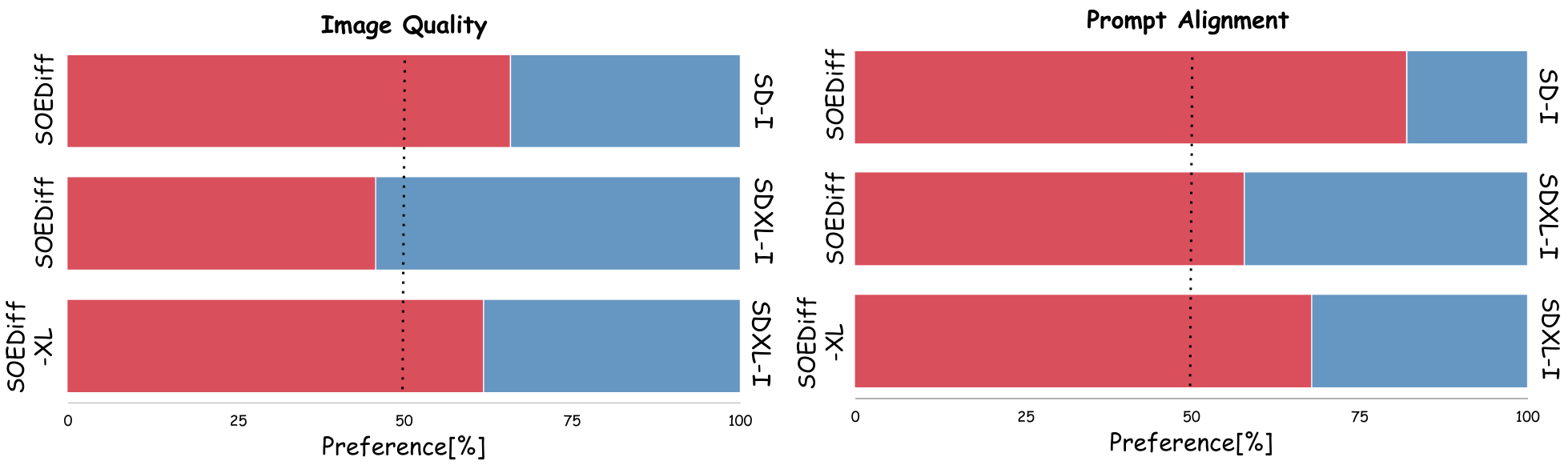}
  \end{center}
  \caption{User preference study. We compare the performance of SOEDiff against baselines SD-I and SDXL-I.}
  \Description{}
  \label{fig:userstudy}
\end{figure}

\noindent\textbf{User Preference Study}. Besides the automated metrics, we also incorporated a user study to align more closely with intuitive human preference.  In the study, we aim to assess both prompt adherence and the overall image. We employed SD-I, SOEDiff, SDXL-I, and SOEDiff-XL as comparison models, keeping the seed fixed to generate 25 sets, totaling 100 images. Our study involved 100 total participants. Participants were tasked with ranking images of small target objects generated by the four models. We present the user study in Fig.~\ref{fig:userstudy} and find that SOEDiff significantly enhances the base model's ability to improve both the image quality of generated objects and their alignment with the associated text.

\section{Conclusion}\label{sec:Conclusion}
In this study, we introduced SOEDiff, a cutting-edge approach designed to significantly enhance the editing of small objects using diffusion models. Central to our methodology is the integration of SO-LoRA, which is meticulously tailored to the small object editing (SOE) task. This innovation allows for precise and efficient adaptation of diffusion models to manage the complexities associated with the editing of small-sized objects within images. Additionally, our adoption of cross-scale score distillation harnesses high-resolution representations from a teacher model to further refine the quality and detail of the generated images. This approach not only improves the fidelity and accuracy of small object edits but also addresses common challenges such as blurriness and text-image mismatches typically seen in standard diffusion model applications.

Our extensive evaluations across standard datasets like MSCOCO and OpenImage validate the effectiveness of SOEDiff, demonstrating substantial improvements over existing models in terms of both CLIP-Score and FID metrics. These results underscore the potential of our method to not only advance the state of the art in small object editing but also to provide a viable solution for applications requiring detailed and precise modifications at a granular level.

\section{Future Work}\label{sec:future work}
Looking forward, the promising results of SOEDiff open avenues for further exploration. First, we aim to extend the capabilities of our model from editing single small-sized objects to editing multiple objects simultaneously. Second, while SOEDiff supports more flexible masks beyond current bounding boxes, developing evaluation metrics for these types of small-sized objects presents challenges due to the intrinsic small size. A potential workaround could involve applying super-resolution techniques prior to evaluation to enhance the measurement of these edits.


\bibliographystyle{ACM-Reference-Format}
\bibliography{sample-base}


\end{document}